\documentclass[journal]{IEEEtran}
\usepackage[utf8]{inputenc}
\usepackage[T1]{fontenc}
\usepackage{cite}
\usepackage{amsmath,amssymb,amsfonts}
\usepackage{graphicx}
\usepackage{textcomp}
\usepackage{xcolor}
\usepackage{booktabs}
\usepackage{colortbl}
\usepackage{multirow}
\usepackage{array}
\usepackage{tabularx}
\usepackage{url}
\usepackage{xspace}
\usepackage[colorlinks=true,linkcolor=blue,citecolor=blue,urlcolor=blue]{hyperref}

\newcommand{\method}{PLAN-S\xspace}

\begin{document}

\title{PLAN-S: Bridging Planning with Latent Style Dynamics for Autonomous Driving World Models}

\author{Xiaoyun~Qiu,
        Jingtao~He,
        Yijie~Chen,       
        Yusong~Huang,
        Haotian~Wang,
        Yixuan~Wang,
        and~Xinhu~Zheng%
\thanks{All authors are with the
  Intelligent Transportation Thrust, Systems Hub, and Center of Seamless Connectivity \& Connected Intelligence,
  The Hong Kong University of Science and Technology (Guangzhou), Guangzhou 511453, China.
  Corresponding author: Xinhu Zheng.}%
}

\markboth{Submitted to IEEE Transactions on Intelligent Transportation Systems}%
{Qiu \MakeLowercase{\textit{et al.}}: PLAN-S}

\maketitle

\begin{abstract}
Latent world models (LWMs) have strengthened end-to-end autonomous driving by forecasting compact scene dynamics for downstream planning. However, existing LWM-based planners usually generate trajectories directly from entangled latent representations. This compact latent-to-planner pathway lacks explicit modeling of risk, drivability, and diverse style preferences, making driving-style dynamics difficult to supervise, inspect, or modulate before a final trajectory is selected. We propose \method (\textbf{PLAN}ning with latent \textbf{S}tyle dynamics), a planner-facing bridge that addresses this compactness-controllability dilemma by decoding a style-conditioned, four-channel semantic cost map from the latent representation. The cost map is conditioned on ego state and driving style and is consumed upstream of the planning decision through two host-side interfaces: attention-level fusion for regression planners and reward-level fusion for anchor-score planners. We validate \method on two architecturally distinct hosts, ResWorld on nuScenes and WoTE on NAVSIM, while keeping the host backbones frozen to isolate the contribution of the proposed bridge. On nuScenes, \method reduces L2 at every horizon over the baseline, with $0.55$\,m average L2 and a $42\%$ relative reduction in the 3\,s collision rate. On NAVSIM, the rule-cost variant reaches $89.4$ Predictive Driver Model Score (PDMS), while the learned cost variant provides complementary gains on baseline-challenging scenes. Ablations show that the cost pathway contributes most directly to safer trajectory selection. Qualitative results further show that \method can produce diverse cost maps, with spatially consistent variations aligned to different driving styles.
\end{abstract}


\section{Introduction}\label{sec:intro}

\IEEEPARstart{R}{ecent} advances in end-to-end (E2E) autonomous driving have demonstrated the potential of learning unified representations that directly connect sensor observations with driving decisions. Along this line, latent world models (LWMs)~\cite{occworld,driveworld,bevworld,resworld} further extend this paradigm from direct decision learning to latent dynamics modeling, where compact latent representations are forecast to capture future scene evolution. By learning planning policies directly on these latent representations, LWM-based planners have demonstrated strong planning capabilities in autonomous driving.

Despite providing a compact and effective pathway, the latent-to-planner paradigm still lacks explicit modeling of key driving properties, including risk, drivability, and diverse style preferences. Existing LWM-based planners, including regression-based methods~\cite{uniad,vad,genad,resworld} and anchor-score-based methods~\cite{li2024hydra,diffusiondrive,wote}, usually generate continuous or discrete trajectories directly from entangled latent representations, where scene dynamics and driving-style dynamics are implicitly coupled~\cite{styledrive,thousandfaces,wang2026drive}. This implicit coupling makes driving-style dynamics difficult to supervise, inspect, or modulate before committing to a final trajectory.

In this paper, we aim to address the compactness-controllability dilemma by explicitly modeling style dynamics on latent representations. We argue that effective latent style dynamics should satisfy three key properties. First, the latent space should be \textit{explicitly controllable}, allowing risk-, drivability-, and route-related preferences to be supervised, visualized, and ablated. Second, the latent style dynamics should be \textit{planning-oriented}, where latent representations are organized as spatial costs rather than optimized for scene reconstruction. Third, it should be \textit{portable across different families} of LWM-based planners, as decoding heads from different paradigms, such as regression-based and anchor-score-based approaches, introduce distinct decision policies through different head couplings.

\begin{figure*}[t]
\centering
\includegraphics[width=\textwidth]{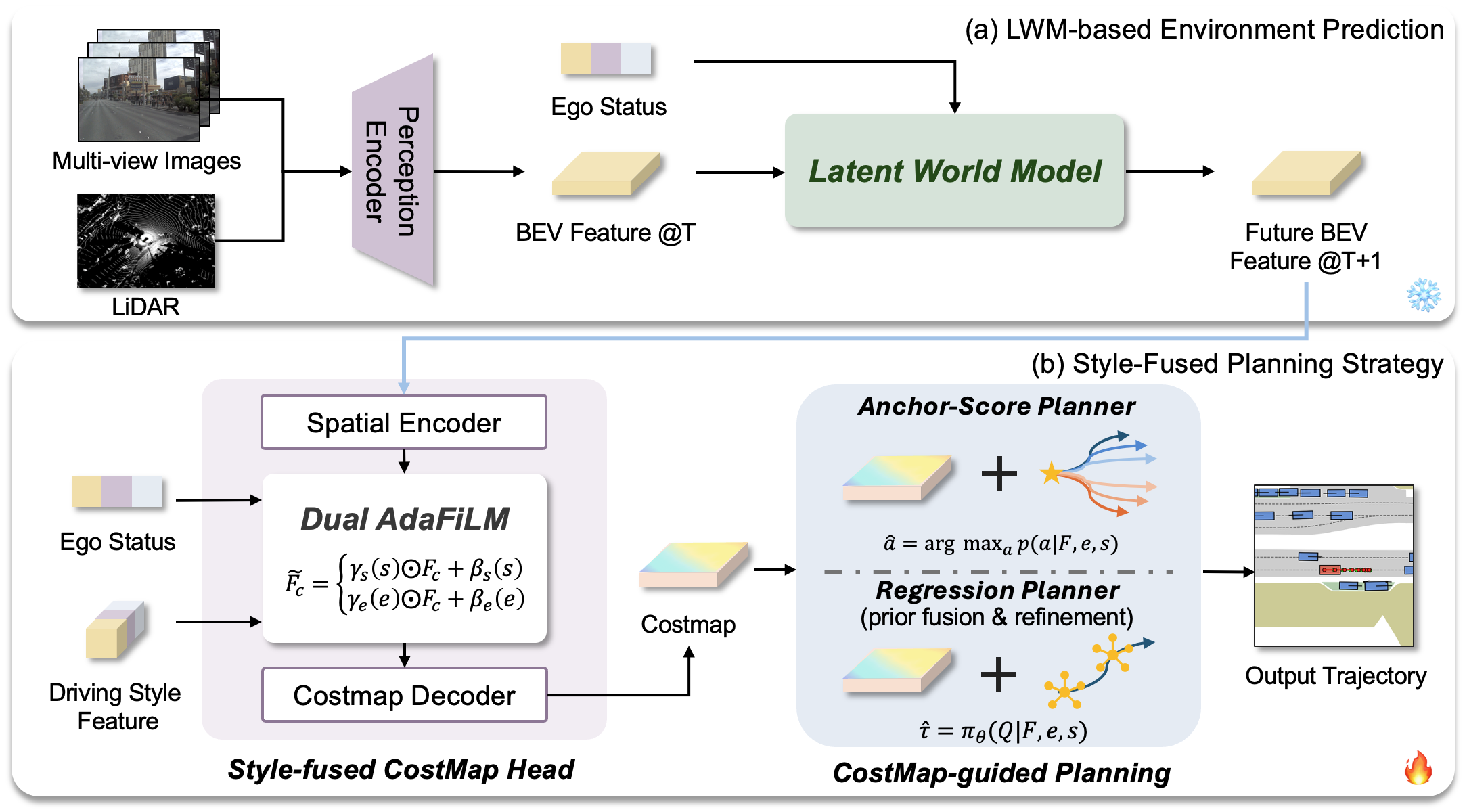}
\caption{Overview of the \method framework. \textbf{(a)} The frozen perception encoder and latent world model produce current and future bird's-eye-view (BEV) features. \textbf{(b)} The trainable cost-map decoder produces a four-channel semantic cost map from the BEV latent, conditioned on ego state and driving style via dual AdaFiLM. The cost map then guides planning through matched coupling interfaces for anchor-score and regression planners.}
\label{fig:framework}
\end{figure*}

We therefore propose \method (\textbf{PLAN}ning with latent \textbf{S}tyle dynamics), a planner-facing bridge between the latent representation and the planning head, as illustrated in Fig.~\ref{fig:framework}. \method explicitly models latent style dynamics as a four-channel semantic cost map, covering dynamic obstacles, off-road regions, static obstacles, and drivability. The cost map is conditioned on the instantaneous ego state and a driving-style code through a dual adaptive feature-wise linear modulation (dual AdaFiLM) mechanism inspired by feature-wise linear modulation (FiLM)~\cite{film}. This design couples scene-level risk with driver-level intent before the final planning decision. It also keeps the intermediate representation inspectable and planning-oriented, rather than treating style as a post hoc adjustment to the output trajectory.

To support different LWM-based planners, \method exposes two host-side interfaces over the same cost-map contract. For regression planners, the cost map is injected through attention-level fusion, where it biases waypoint queries and guides spatial refinement. For anchor-score planners, the same cost map is injected through reward-level fusion, where sampled costs along candidate anchors adjust the native anchor scores. Both interfaces use the cost map before the final trajectory is selected, but they respect the decision form of each planner family. This makes the proposed latent style dynamics portable across regression-based and anchor-score-based planners.

We validate this design on two architecturally distinct hosts: ResWorld~\cite{resworld} with nuScenes~\cite{nuscenes} for the regression case and WoTE~\cite{wote} with NAVSIM~\cite{navsim} for the anchor-score case. Both instantiations leave the host backbone unchanged and use no weight sharing across hosts. This isolates the effect of the proposed latent style dynamics. The main contributions are summarized as follows:
\begin{itemize}
\item We introduce a four-channel semantic cost map that explicitly models risk, drivability, and route-related preferences as planning-oriented latent style dynamics.
\item We instantiate the same cost-map contract in regression-based and anchor-score-based LWM planners through attention-level fusion and reward-level fusion, without modifying the host backbone.
\item We condition the planner-facing intermediate on ego state and driving style through separate AdaFiLM pathways, enabling style-dependent cost-map modulation before final trajectory selection.
\end{itemize}

\section{Related Work}\label{sec:related}

\subsection{Latent World Models in E2E Autonomous Driving}
Latent world models compress the driving scene into a compact feature space for predicting future scene evolution and supporting downstream planning. In E2E autonomous driving, this latent feature can be organized in different forms, including tokenized visual latents, occupancy latents, and bird's-eye-view (BEV) latents. MILE~\cite{mile} learns a discrete latent dynamics model via a vector-quantized variational autoencoder (VQ-VAE) for imagination-based planning, establishing that world-model rollouts in a compact latent can replace explicit perception-then-prediction pipelines. OccWorld~\cite{occworld} autoregresses 3D occupancy tokens via a generative pre-trained transformer (GPT)-style generator. DriveWorld~\cite{driveworld} couples a state-space memory with a dynamic memory bank to forecast 4D scene state. BEVWorld~\cite{bevworld} predicts future BEV latents via a latent sequence diffusion model~\cite{dit} conditioned on action tokens, avoiding autoregressive error accumulation. ResWorld~\cite{resworld} models the residual between successive BEV states and refines candidate trajectories against the predicted future BEV using a transformer over sparse scene queries. WoTE~\cite{wote} rolls anchor-conditioned future BEV latents for each candidate trajectory and scores them with a learned reward model combining imitation and simulation rewards. Beyond BEV-anchored models, GAIA-1~\cite{gaia} generates future driving scenes autoregressively in a tokenized video latent space, and DriveWM~\cite{drivewm} forecasts multiview images and uses the predicted futures to guide planning.

Among these latent forms, BEV has become a common choice for E2E autonomous driving models because it preserves metric spatial layout in a top-down view. The lift-splat formulation of LSS~\cite{lss}, followed by BEVFormer~\cite{bevformer} and BEVDepth~\cite{bevdepth}, shows how multi-camera observations can be transformed into a spatial grid where lanes, agents, and drivable regions are geometrically aligned. These spatial latents provide a suitable basis for further planning-oriented modeling, especially when the goal is to preserve spatial awareness while keeping the representation compact.

However, existing BEV-latent planners usually consume the latent as an implicit feature tensor, so the spatial information is not explicitly represented as planner-facing risk, drivability, or preference semantics. They also rarely model driving style within this spatial intermediate, leaving style-dependent planning preferences difficult to supervise, inspect, or modulate before trajectory selection.

\subsection{Planning-Oriented Autonomous Driving and Cost/Reward-Guided Planning}
Modern E2E planners can be broadly divided into regression-based planners and anchor-score planners according to the form of the planning output. Regression-based planners directly generate continuous trajectories from BEV-conditioned queries or latent features, whereas anchor-score planners restrict the action space to a discrete set of trajectory anchors and select among them by scoring candidates. Representative regression-based methods include UniAD~\cite{uniad} and VAD~\cite{vad}, which use transformer decoders with command-conditioned planning queries; TCP~\cite{tcp}, which combines trajectory prediction with direct control prediction; GenAD~\cite{genad} and GoalFlow~\cite{goalflow}, which formulate trajectory generation through generative modeling and flow matching; SparseDrive~\cite{sparsedrive}, which uses sparse scene representations with parallel task heads; and ResWorld~\cite{resworld}, which refines waypoint queries against predicted future BEV features. Representative anchor-score methods include Hydra-MDP~\cite{li2024hydra}, which learns multi-head trajectory scoring with rule-based rewards; DiffusionDrive~\cite{diffusiondrive}, which refines anchor-initialized trajectories through truncated diffusion; and WoTE~\cite{wote}, which scores anchors by rolling future BEV latents and evaluating them with a learned reward model. This anchor-score paradigm is also well aligned with closed-loop benchmarks such as NAVSIM~\cite{navsim} and the Predictive Driver Model (PDM) family of planners~\cite{pdm}.

Cost and reward signals have been introduced into both planning paradigms to make trajectory selection more planning-aware. Occupancy prediction provides spatial cues by forecasting where agents or obstacles may appear, as in UniAD~\cite{uniad} and OccWorld~\cite{occworld}, but it remains primarily a perception-side prediction target. Planning-side cost and reward formulations more directly assign desirability to candidate actions. MP3~\cite{mp3} plans over a learned cost function derived from predicted occupancy and flow, NMP~\cite{zeng2019end} predicts a learned cost volume and scores trajectories with a max-margin objective, and ST-P3~\cite{stp3} combines hand-crafted costs from BEV segmentation with a learned cost for candidate trajectory selection. In anchor-score frameworks, rule-based rewards~\cite{li2024hydra} and learned score functions~\cite{wote,diffusiondrive} play a similar role by mapping scene features to per-anchor desirability.

These methods demonstrate that cost and reward cues are useful for planning, but their spatial representations are usually co-designed with a specific planner family and a specific consumption mechanism. As a result, existing planning-oriented methods still lack a cross-family, explicit spatial representation that can serve as a common interface for both regression-based and anchor-score LWM planners.

\subsection{Style-Aware and Personalized Driving}
Driving style and personalization in autonomous driving have received growing attention. StyleDrive~\cite{styledrive} contributes a large-scale real-world dataset with aggressive, normal, and conservative labels and proposes the style-matching Predictive Driver Model Score (SM-PDMS) metric; rather than introducing a new architecture, it evaluates simple style-conditioned variants of existing E2E autonomous driving models. \emph{Driving with a Thousand Faces}~\cite{thousandfaces} introduces a personalized E2E autonomous driving benchmark together with a style-reward-model-based lightweight fine-tuning framework that adapts a base planner to individual driving preferences. \emph{Drive My Way}~\cite{wang2026drive} conditions a vision-language-action policy on learned user embeddings derived from a personalized driving dataset and applies reinforcement fine-tuning with weighted safety/comfort/efficiency rewards to adapt to natural-language style instructions.
Earlier variational autoencoder (VAE)-based methods~\cite{vaegru} infer latent style vectors of surrounding drivers from observed trajectories in an unsupervised manner and feed these vectors into the state representation of the reinforcement learning (RL) policy, enabling the ego vehicle to condition decisions on the inferred styles of other traffic participants.
Anchor-based planners such as \emph{DiffusionDrive}~\cite{diffusiondrive} and VADv2~\cite{vadv2} partition the trajectory space via pre-clustered anchors, but anchor membership is learned from aggregated demonstrations without an explicit style-control channel. Diversity therefore emerges from the anchor distribution rather than from user-specified intent.

Across this body of work, style information is mainly applied at the output stage through trajectory selection, reward weighting, or final policy conditioning. This makes it difficult to inspect how style changes planning-relevant spatial semantics before the final decision is made.

\section{Methodology}\label{sec:method}

As illustrated in Fig.~\ref{fig:framework}, our framework inserts a style-conditioned cost-map interface between the BEV latent representation and the planning head. The shared decoder converts the implicit BEV latent into a four-channel semantic cost map that explicitly represents dynamic obstacles, off-road regions, static obstacles, and drivability. This cost map is conditioned on ego state and driving style, so style-dependent planning preferences are encoded before trajectory selection. To support different planner families, the same cost-map representation is coupled to regression planners through attention-level fusion and to anchor-score planners through reward-level fusion, while the host backbone remains unchanged.

\subsection{Problem Formulation and Interface Design}\label{sec:problem}
Let $F \in \mathbb{R}^{C \times H \times W}$ denote the BEV latent feature produced by an LWM backbone at a given timestep, where $C$ is the channel dimension and $H \times W$ is the spatial extent of the BEV grid. Let $\tau = \{(x_t, y_t)\}_{t=1}^{T}$ denote the ego trajectory over a planning horizon of $T$ waypoints, $e \in \mathbb{R}^{d_e}$ the instantaneous ego state (velocity, yaw rate, acceleration), $s \in \mathbb{R}^{d_s}$ the driving-style code, and $\mathrm{cmd}$ the high-level navigation command.

An LWM planner is a function $\pi_\theta$ that maps BEV features and auxiliary conditioning to a trajectory: $\tau = \pi_\theta(F, e, \mathrm{cmd})$. For regression heads, $\pi_\theta$ outputs $\tau$ directly. For anchor-score heads, it produces a distribution over a fixed anchor set $\mathcal{A} = \{\tau^{(a)}\}_{a=1}^{|\mathcal{A}|}$ via scores $r \in \mathbb{R}^{|\mathcal{A}|}$ and selects $\tau = \tau^{(a^\star)}$ with $a^\star = \arg\max_a r_a$. \method inserts a cost-map interface $\mathcal{C}$ between the backbone and the planner:
\begin{equation}
M = \mathcal{C}(F, s, e) \in \mathbb{R}^{K \times H \times W},
\end{equation}
where $K$ denotes the number of semantic cost channels, set to $K=4$ in our implementation. A host-specific interface $\tilde{\pi}_\theta$ then consumes both $F$ and $M$: $\tau = \tilde{\pi}_\theta(F, M, e, \mathrm{cmd})$. In this work, ``pluggability'' means that the same cost-map module $\mathcal{C}$ exposes a shared input--output contract across hosts and can be coupled to different planner families without altering the host perception backbone. The wrapper $\tilde{\pi}_\theta$ may remain host-specific, and different instantiations may use independent training, different spatial resolutions, and different auxiliary supervision, provided that the cost-map representation retains the same channel semantics.

\label{sec:interface}
The module exposes a minimal interface to each host. The input consists of a BEV feature map $F$, an ego state vector $e$, and a driving-style code $s$. The output is a four-channel cost map $M \in \mathbb{R}^{K \times H \times W}$ aligned to the BEV grid of the host, together with a per-channel logit map $\hat{M}$ for supervision. The supervised loss $\mathcal{L}_{\mathrm{cost}}$ is an optional training signal that can be warmed up independently of the planning loss of the host. Hosts may differ in BEV resolution, auxiliary target construction, and consumption mechanism; the invariant component is the four-channel semantic contract of $M$.

\subsection{Style-Conditioned Cost-Map Decoder}\label{sec:sfch}
The shared module comprises three components: a driving-style encoder, a dual AdaFiLM modulation stage, and a four-channel cost decoder. The same module design is used for both regression-planner and anchor-score-planner instantiations, while feature dimensions and grid size are matched to the native BEV representation of each host.

\subsubsection{Driving Style Encoder}
The interface only requires a driving-style code $s$; its encoder can be matched to the data protocol and host planner without changing the subsequent cost-map decoder. For the regression-planner instantiation, $s$ is derived from an ego kinematic history $\{u_{t-l}\}_{l=0}^{L-1}$ by a compact recurrent encoder, where $u_{t-l}=(x_{t-l}, y_{t-l}, \psi_{t-l}, v_{t-l}, a_{t-l})$ contains position, heading, speed, and acceleration: $s = \mathrm{GRU}(\{u_{t-l}\}_{l=0}^{L-1}) \in \mathbb{R}^{d_s}$, with GRU denoting gated recurrent unit. We use a 64-dimensional continuous embedding to preserve richer temporal preference information for trajectory regression. For the anchor-score-planner instantiation, we use a deterministic two-dimensional style score $s \in [0, 1]^2$ derived from ego kinematics, with the two dimensions corresponding to longitudinal and lateral aggressiveness. Both representations are mapped to the same subsequent modulation stage, so the instantiation-specific encoder affects only how the style code is obtained, not the semantic contract of the cost map.

\subsubsection{Dual AdaFiLM Modulation}
Classical FiLM~\cite{film} modulates feature maps via an affine transformation $(\gamma, \beta)$ predicted from a conditioning signal. Dual AdaFiLM extends this with a channel-split design. The motivation is that the two conditioning signals differ in nature: the style code $s$ is internal and preference-like, whereas the ego state $e$ is an observed vehicle state. Mixing them under a single affine map can entangle the two effects and make style-specific behavior harder to inspect. We partition the conditioned channel indices into two disjoint groups $\mathcal{I} = \mathcal{I}_s \cup \mathcal{I}_e$ with $|\mathcal{I}_s| = |\mathcal{I}_e|$, and modulate each group independently:
\begin{equation}
\tilde{F}_c = \begin{cases} \gamma_{s,c}(s) F_c + \beta_{s,c}(s), & c \in \mathcal{I}_s, \\ \gamma_{e,c}(e) F_c + \beta_{e,c}(e), & c \in \mathcal{I}_e, \end{cases}
\end{equation}
where $F_c$ and $\tilde{F}_c$ are the $c$-th input and modulated BEV feature channels. The per-channel affine parameters $(\gamma_s, \beta_s)$ and $(\gamma_e, \beta_e)$ are produced by two independent two-layer multilayer perceptrons (MLPs) and broadcast over the spatial dimensions. The partition is fixed and non-learned; all channel mixing occurs in the subsequent decoder, which is free to recombine the two groups. This design encourages separation between style-driven and ego-state-driven modulation at the point of conditioning while still permitting flexible downstream fusion.

\subsubsection{Four-Channel Semantic Cost Decoder}
The modulated BEV feature $\tilde{F}$ is decoded by a lightweight convolutional head into a four-channel cost map $M$. The four channels decompose planning-relevant information per cell: a \emph{dynamic} channel for moving agents (vehicles, pedestrians, cyclists), an \emph{off-road} channel for regions outside the legal drivable surface, a \emph{static} channel for stationary obstacles (barriers, curbs, cones), and a \emph{drivability} channel that encodes a positive (inverse-cost) signal for preferred lanes and route-aligned cells. This channel design separates hazards that should generally be avoided from route-aligned cells that should be rewarded, making the downstream fusion weights physically interpretable. When an auxiliary target is available, each channel is supervised by a binary cross-entropy (BCE) loss:
\begin{equation}
\mathcal{L}_{\mathrm{cost}} = \sum_{k=1}^{K} \mathrm{BCE}\bigl(\sigma(\hat{M}_k),\, M_k^{\mathrm{gt}}\bigr),
\end{equation}
where $\sigma(\cdot)$ is the sigmoid function, $\hat{M}$ denotes the pre-sigmoid logits of the cost map, $M^{\mathrm{gt}}$ is the channel-wise supervision target, and $\mathrm{BCE}(\cdot,\cdot)$ is averaged over BEV cells. The construction of $M^{\mathrm{gt}}$ is host-dependent because different planners expose different map annotations, semantic outputs, and BEV resolutions. This loss is therefore used as an auxiliary interface for learning a spatially coherent cost map, while the planning objective remains responsible for optimizing the final trajectory decision.

\subsection{Cost-Map-Guided Planning Interfaces}\label{sec:coupling}

\subsubsection{Attention-Level Coupling with Regression Planners}\label{sec:branch1}
Regression-based planners refine a set of waypoint queries $Q \in \mathbb{R}^{T \times d}$ through iterative attention over the BEV latent. Our coupling introduces two insertion points for this planner family.

The \emph{upstream} interface performs cost-conditioned prior fusion. Before any deformable refinement, the waypoint queries $Q$ attend into a flattened, position-encoded cost map $M$ via a cross-attention layer:
\begin{equation}
Q' = Q + \mathrm{CrossAttn}\bigl(Q,\, \phi_M(\mathrm{flatten}(M)) + \mathrm{PE}\bigr),
\end{equation}
where $\phi_M$ is a linear projection that maps flattened cost-map tokens to the query dimension and $\mathrm{PE}$ denotes positional encoding on the BEV grid. The resulting cost-conditioned prior $Q'$ initializes the deformable refinement stack. This ensures that the initial trajectory hypothesis is already informed by the spatial structure of the cost map, so the cost map serves not merely as a re-scoring mechanism but directly shapes the prior that the planner will subsequently refine.

The \emph{downstream} interface performs cost-guided deformable attention. The deformable attention operator is augmented with a per-head gate $g \in [0, 1]^{H_a}$, where $H_a$ is the number of attention heads. Each head $h$ is assigned a primary cost channel $k(h)$, and the contribution to the residual update is scaled by $g_h \cdot \sigma(\hat{M}_{k(h)})$ at the sampled location, turning the cost map into a spatial attenuator for the refinement signal.

\subsubsection{Reward-Level Coupling with Anchor-Score Planners}\label{sec:branch2}
Anchor-score planners evaluate a discrete anchor set $\mathcal{A}$ by assigning a reward or score to each candidate trajectory $\tau^{(a)} \in \mathcal{A}$. This is the native decision mechanism of this planner family, and we use it without modifying the host planner.

For anchor-score planners, the cost sampler bilinearly samples each channel of the cost logit map $\hat{M}$ along the waypoints of each anchor and aggregates across the $K$ channels:
\begin{equation}
c^{(a)} = \frac{1}{T} \sum_{t=1}^{T} \sum_{k=1}^{K} w_k\,\hat{M}_k\bigl(\tau^{(a)}_t\bigr),
\end{equation}
where $\hat{M}_k(\tau^{(a)}_t)$ is obtained by bilinear sampling at waypoint $\tau^{(a)}_t$, $w_k{=}{+1}$ for the three obstacle channels (dynamic, off-road, static), and $w_k{=}{-1}$ for the drivability channel, so that high drivability reduces the anchor penalty. This produces a scalar per-anchor cost $c^{(a)}$ for all $|\mathcal{A}|{=}256$ anchors through batched bilinear sampling. We then combine these per-anchor costs with the native anchor reward $r \in \mathbb{R}^{|\mathcal{A}|}$ in log-space, matching the multiplicative structure of the underlying anchor likelihood:
\begin{equation}
r_{\mathrm{fused}}^{(a)} = r^{(a)} + \lambda_{\mathrm{cost}} \cdot \log\bigl(1 - \sigma(c^{(a)})\bigr),
\end{equation}
where $\lambda_{\mathrm{cost}}$ is a scalar fusion weight. The added term is non-positive and monotonically decreases as the sampled cost increases, so it acts as a cost-dependent penalty on the native anchor reward. In implementation, the logarithm is evaluated with a small numerical clamp to avoid the singular point when $\sigma(c^{(a)})$ approaches one. This formulation is compatible with the probabilistic composition of the existing reward and remains stable in the low-cost regime. It provides a conservative way to insert a spatial prior into an existing anchor scorer. The future latent decoder, semantic head, and anchor vocabulary of the host remain unmodified; \method contributes only the shared cost-map decoder and the scoring and fusion interfaces described above.

\subsubsection{Analysis of the Two Interfaces}\label{sec:comparison}
The two interfaces target different planner families but follow the same design principle: the cost map is introduced before the planner finalizes its decision variable and is then used again to refine that decision with spatial evidence. In regression planners, the decision variable is the waypoint query; in anchor-score planners, it is the score assigned to each trajectory anchor. This correspondence yields a consistent prior-plus-refinement pattern across both families. The explicit four-channel cost map also makes the inserted spatial evidence inspectable: each channel has a defined semantic role, and each fusion step can be traced to locations and channels on the BEV grid. The design therefore turns an otherwise compact latent-to-planner pathway into a planner-facing representation that is more explicit, reusable, and easier to interpret across different planning heads.

\section{Experiments}\label{sec:exp}
This section evaluates \method on two standard autonomous driving benchmarks and examines its planning performance. We begin by summarizing the datasets, metrics, baselines, and implementation details in Sec.~\ref{sec:exp_setup}. We then report the main results and ablation studies for the regression-planner instantiation on nuScenes in Sec.~\ref{sec:b1_main}, followed by the anchor-score instantiation on NAVSIM in Sec.~\ref{sec:b2_main}. The section concludes with qualitative analysis of the learned cost maps and style-conditioned behavior in Sec.~\ref{sec:qual_costmap}.

\subsection{Experimental Setup}\label{sec:exp_setup}

\textbf{Datasets and metrics.} We evaluate \method on two benchmarks matched to the two planner families studied in this paper. The ResWorld~\cite{resworld} instantiation is evaluated on nuScenes~\cite{nuscenes}, the standard open-loop benchmark for E2E autonomous driving planning with camera and multi-modal inputs, to validate the regression-planner interface. We follow the official training and validation splits (700 and 150 scenes) and report L2 displacement error and collision rate at 1\,s, 2\,s, and 3\,s horizons. The WoTE~\cite{wote} instantiation is evaluated on NAVSIM~\cite{navsim}, a closed-loop reactive-simulation benchmark built on nuPlan scenes, to validate the anchor-score interface. We use \texttt{navtrain} for training and \texttt{navtest} for evaluation, and report the Predictive Driver Model Score (PDMS), which multiplicatively aggregates No-Collision (NC), Drivable-Area Compliance (DAC), Time-to-Collision (TTC), Comfort (Cmf), and Ego-Progress (EP). We do not force both hosts into both open-loop and closed-loop protocols because the two benchmarks expose different native planner interfaces, annotations, and evaluation contracts; using each host in its standard protocol provides the cleanest controlled test of the corresponding coupling interface.

\begin{table*}[!t]
\centering
\caption{Regression-planner results on nuScenes \texttt{val}. $\diamondsuit$ denotes ego-status inputs following BEVPlanner++~\cite{bevplannerpp}; $\ast$ denotes officially released models. L2 is per-second endpoint displacement with Avg as the simple mean of the three horizons; CR is reported as the percentage of trajectories that collide within the corresponding horizon.}
\label{tab:b1_main}
\footnotesize
\setlength{\tabcolsep}{3pt}
\renewcommand{\arraystretch}{1.05}
\begin{tabularx}{\textwidth}{@{}>{\raggedright\arraybackslash}p{0.15\textwidth}>{\raggedright\arraybackslash}p{0.23\textwidth}*{8}{>{\centering\arraybackslash}X}@{}}
\toprule
\multirow{2}{*}{Method} & \multirow{2}{*}{Auxiliary Task} & \multicolumn{4}{c}{L2 (m)$\downarrow$} & \multicolumn{4}{c}{Collision Rate (\%)$\downarrow$} \\
\cmidrule(lr){3-6}\cmidrule(lr){7-10}
 & & 1s & 2s & 3s & Avg & 1s & 2s & 3s & Avg \\
\midrule
ST-P3~\cite{stp3}                  & Det\&Map                     & $1.72$ & $3.26$ & $4.86$ & $3.28$ & $0.44$ & $1.08$ & $3.01$ & $1.51$ \\
UniAD~\cite{uniad}                 & Det\&Track\&Map\&Motion\&Occ & $0.48$ & $0.96$ & $1.65$ & $1.03$ & $0.05$ & $0.17$ & $0.71$ & $0.31$ \\
OccNet~\cite{occnet}               & Det\&Map\&Occ                & $1.29$ & $2.13$ & $2.99$ & $2.14$ & $0.21$ & $0.59$ & $1.37$ & $0.72$ \\
PARA-Drive~\cite{paradrive}        & Det\&Track\&Map\&Motion\&Occ & $0.40$ & $0.77$ & $1.31$ & $0.83$ & $0.07$ & $0.25$ & $0.60$ & $0.30$ \\
GenAD~\cite{genad}                 & Det\&Map\&Motion             & $0.36$ & $0.83$ & $1.55$ & $0.91$ & $0.06$ & $0.23$ & $1.00$ & $0.43$ \\
SSR$\ast$~\cite{ssr}               & None                         & $0.25$ & $0.64$ & $1.33$ & $0.74$ & $0.08$ & $0.12$ & $0.72$ & $0.31$ \\
ResWorld~\cite{resworld}           & None                         & $0.22$ & $0.56$ & $1.17$ & $0.65$ & $0.02$ & $0.04$ & $0.64$ & $0.23$ \\
ResWorld$\diamondsuit$~\cite{resworld} & None & $0.19$ & $0.50$ & $1.08$ & $0.59$ & $0.02$ & $0.06$ & $0.43$ & $0.17$ \\
\midrule
\rowcolor{gray!15}
\textbf{\method}               & Cost map & $\mathbf{0.17}$ & $\mathbf{0.46}$ & $\mathbf{1.01}$ & $\mathbf{0.55}$ & $\mathbf{0.02}$ & $\mathbf{0.06}$ & $\mathbf{0.25}$ & $\mathbf{0.11}$ \\
\bottomrule
\end{tabularx}
\end{table*}
\textbf{Baselines.} On nuScenes, we compare with explicit-intermediate methods (ST-P3~\cite{stp3}, UniAD~\cite{uniad}, OccNet~\cite{occnet}, PARA-Drive~\cite{paradrive}), structured or generative regression planners (GenAD~\cite{genad}, SSR~\cite{ssr}), and the LWM regression host ResWorld~\cite{resworld}. On NAVSIM, we compare with representative E2E planners (VADv2~\cite{vadv2}, UniAD~\cite{uniad}, LTF~\cite{transfuser}, PARA-Drive~\cite{paradrive}, LAW~\cite{law}, TransFuser~\cite{transfuser}, DRAMA~\cite{drama}) and anchor-score or reward-guided planners (DiffusionDrive~\cite{diffusiondrive}, Hydra-MDP~\cite{li2024hydra}, WoTE~\cite{wote}). ResWorld and WoTE are the direct hosts and are reproduced for controlled comparisons; the other methods provide benchmark context.

\label{sec:impl}
\label{sec:training}
\textbf{Implementation.} Both instantiations are implemented in PyTorch, trained on $8\times$ NVIDIA H100 graphics processing units (GPUs), and keep the host backbones frozen. The cost-map grids follow the native BEV resolutions of the hosts, namely $200\times200$ for ResWorld and $64\times64$ for WoTE. During training, style dropout replaces $s$ with the corresponding neutral style code with probability $p_{\mathrm{drop}}=0.1$. The two instantiations are trained independently without weight transfer.

For the regression-planner instantiation, we use the GeoBEV~\cite{zhang2025geobev} backbone of ResWorld and train the added modules for 18 epochs with a batch size of 16. We use AdamW~\cite{adamw} with an initial learning rate of $1\!\times\!10^{-4}$ and a weight decay of $0.01$. The optimization combines the cost-map loss $\mathcal{L}_{\mathrm{cost}}$ and the planning loss $\mathcal{L}_{\mathrm{plan}}$. The cost-map target $M^{\mathrm{gt}}$ is pre-computed offline from nuScenes high-definition (HD) map and annotation data. The dynamic and static channels are rasterized from agent bounding boxes partitioned by motion status, the off-road channel is derived from the complement of the annotated drivable area, and the drivability channel is derived from route-aligned lane polygons. All targets are projected into the host BEV coordinate frame before training.

For the anchor-score instantiation, we use the TransFuser~\cite{transfuser} backbone and train the added modules for 30 epochs with a batch size of 64. We use AdamW with an initial learning rate of $1\!\times\!10^{-4}$ and a weight decay of $1\!\times\!10^{-4}$. The optimization combines the cost-map BCE loss and the anchor-planning loss over the fused reward. Since dense cost-map ground truth is unavailable at the target resolution, $M^{\mathrm{gt}}$ is constructed online from the semantic mask predicted by the frozen WoTE semantic head through Gaussian smoothing. This auxiliary target regularizes the learned cost map toward the spatial semantics of the host, and its empirical effect is analyzed in Sec.~\ref{sec:b2_abl}.

\begin{table}[t]
\centering
\caption{Module ablation on nuScenes \texttt{val}. Cost: cost-map module; AdaFiLM: dual AdaFiLM conditioning. L2 Avg and collision rate (CR) Avg are averaged over the three horizons (1\,s, 2\,s, 3\,s) following Table~\ref{tab:b1_main}.}
\label{tab:b1_abl_module}
\small
\setlength{\tabcolsep}{6pt}
\renewcommand{\arraystretch}{1.05}
\begin{tabular*}{\columnwidth}{@{\hspace{3pt}\extracolsep{\fill}} l c c c c @{\hspace{3pt}}}
\toprule
Method & Cost & AdaFiLM & L2 Avg$\downarrow$ & CR Avg (\%)$\downarrow$ \\
\midrule
$M_1$               & --           & --           & $0.551$          & $0.157$ \\
$M_2$               & $\checkmark$ & --           & $0.549$          & $0.137$ \\
\midrule
\textbf{\method}    & $\checkmark$ & $\checkmark$ & $\mathbf{0.545}$ & $\mathbf{0.111}$ \\
\bottomrule
\end{tabular*}
\end{table}

\begin{table}[t]
\centering
\caption{Coupling-interface ablation on nuScenes \texttt{val}. Up: upstream prior-cost fusion; Down: downstream cost-gated refinement. L2 Avg and collision rate (CR) Avg are averaged over the three horizons (1\,s, 2\,s, 3\,s) following Table~\ref{tab:b1_main}.}
\label{tab:b1_abl_interface}
\small
\setlength{\tabcolsep}{6pt}
\renewcommand{\arraystretch}{1.05}
\begin{tabular*}{\columnwidth}{@{\hspace{3pt}\extracolsep{\fill}} l c c c c @{\hspace{3pt}}}
\toprule
Method & Up & Down & L2 Avg$\downarrow$ & CR Avg (\%)$\downarrow$ \\
\midrule
$M_1$               & --           & $\checkmark$ & $\mathbf{0.541}$ & $0.137$ \\
$M_2$               & $\checkmark$ & --           & $0.557$          & $0.144$ \\
\midrule
\textbf{\method}    & $\checkmark$ & $\checkmark$ & $0.545$ & $\mathbf{0.111}$ \\
\bottomrule
\end{tabular*}
\end{table}

\begin{table}[t]
\centering
\caption{Inference time on a single NVIDIA H100 GPU.}
\label{tab:inference}
\small
\setlength{\tabcolsep}{3pt}
\renewcommand{\arraystretch}{1.05}
\begin{tabular*}{\columnwidth}{@{\hspace{3pt}\extracolsep{\fill}} l c c c @{\hspace{3pt}}}
\toprule
Method & Parameters (M)& Latency (ms)$\downarrow$ & FPS$\uparrow$ \\
\midrule
ResWorld baseline & 81.16 & 64.8 & 15.4 \\
\method & 81.41 & \textbf{59.0} & \textbf{17.0} \\
\bottomrule
\end{tabular*}
\end{table}
\subsection{Regression-Planner Results on nuScenes}\label{sec:b1_main}
\textit{(1) Main results.}
Table~\ref{tab:b1_main} reports the main regression-planner results on the nuScenes \texttt{val} split. Compared with the ResWorld$\diamondsuit$ baseline, \method reduces L2 at all horizons and lowers the average L2 from $0.59$\,m to $0.55$\,m. The larger change appears in collision rate (CR). CR@3s decreases from $0.43\%$ to $0.25\%$, and CR Avg decreases from $0.17\%$ to $0.11\%$. This pattern is consistent with the role of the cost map. L2 mainly rewards proximity to the logged ego trajectory, so it is less sensitive to spatial risk when the predicted trajectory remains close to the human-driven path. CR is more sensitive to risky tail cases. The explicit cost map gives the planner a spatial penalty in these cases before the final trajectory is selected.

\textit{(2) Ablation study.}\label{sec:b1_abl}
The ablation study separates the cost-map module from the planner-coupling interface. Table~\ref{tab:b1_abl_module} tests the cost-map module and dual AdaFiLM conditioning. Table~\ref{tab:b1_abl_interface} tests upstream prior-cost fusion and downstream cost-gated refinement. All variants follow the same evaluation protocol as the main results.

\textit{Module ablation (part a).} Table~\ref{tab:b1_abl_module} shows the effect of each added component. Without the cost-map module, $M_1$ obtains $0.551$\,m L2 Avg and $0.157\%$ CR Avg. Adding the cost map without AdaFiLM reduces CR Avg to $0.137\%$. The full model further reduces it to $0.111\%$. The L2 changes are much smaller, from $0.551$\,m to $0.545$\,m. Thus, the cost-map module mainly affects collision reduction rather than average trajectory displacement. AdaFiLM gives an additional CR gain in this table, and its main role is to make the cost map style-conditioned, which is examined qualitatively in Sec.~\ref{sec:qual_style}.

\textit{Coupling-interface ablation (part b).} Table~\ref{tab:b1_abl_interface} shows that neither interface alone matches the full design. Downstream-only refinement gives the best L2 Avg ($0.541$\,m), but its CR Avg remains $0.137\%$. Upstream-only fusion gives $0.144\%$ CR Avg. Using both interfaces lowers CR Avg to $0.111\%$ while keeping L2 close to the best variant. This supports the prior-plus-refinement design. The upstream interface biases the initial waypoint prior toward lower-cost regions, while the downstream interface applies the same cost evidence during trajectory refinement. The two stages are most useful for collision reduction when they are used together.

\textit{(3) Inference time.}\label{sec:inference}
Table~\ref{tab:inference} reports inference latency on a single NVIDIA H100 80\,GB GPU. The cost-map decoder adds only $0.25$ million parameters ($+0.3\%$) to the ResWorld host. \method runs at $59.0$\,ms, corresponding to $17.0$ frames per second (FPS), while the ResWorld baseline runs at $64.8$\,ms and $15.4$\,FPS. These results show that the proposed bridge improves planning behavior without reducing inference efficiency.

\subsection{Anchor-Score Results on NAVSIM}\label{sec:b2_main}
\textit{(1) Main results.}
Table~\ref{tab:b2_main} reports the main anchor-score results on NAVSIM \texttt{navtest}, including the baseline set from the WoTE evaluation~\cite{wote}. \method (rule) uses a hand-designed cost for log-additive reward fusion. \method (learned) uses the learned style-conditioned cost-map bridge.

Both variants improve over WoTE. The rule variant reaches $89.4$ PDMS, and the learned variant reaches $89.1$ PDMS. The gain is mainly from TTC, which increases from $94.9$ to $97.7$ and $97.5$, respectively. Other sub-metrics are nearly unchanged or slightly lower. This pattern is consistent with anchor-score fusion: the cost term mainly changes the ranking of anchors that pass close to high-cost regions, so its clearest effect appears in TTC. The rule cost is the best aggregate variant on \texttt{navtest}. The learned cost is slightly weaker overall, so we further examine where it helps through the scene-level analysis.

\begin{table*}[t]
\centering
\caption{Anchor-score results on NAVSIM \texttt{navtest}. C: camera; L: LiDAR; Ego: ego-state only. Bold denotes the best non-human result or ties.}
\label{tab:b2_main}
\setlength{\tabcolsep}{8pt}
\small
\renewcommand{\arraystretch}{1.05}
\begin{tabular}{l c c c c c c c}
\toprule
Method & Input & NC$\uparrow$ & DAC$\uparrow$ & EP$\uparrow$ & TTC$\uparrow$ & Cmf$\uparrow$ & PDMS$\uparrow$ \\
\midrule
Human                    & --  & 100.0 & 100.0 & 87.5 & 100.0 & 99.9  & 94.8 \\
Constant Velocity        & --  & 69.9  & 58.8  & 49.3 & 49.3  & \textbf{100.0} & 21.6 \\
Ego Status MLP           & Ego & 93.0  & 77.3  & 62.8 & 83.6  & \textbf{100.0} & 65.6 \\
VADv2~\cite{vadv2}       & C   & 97.9  & 91.7  & 77.6 & 92.9  & \textbf{100.0} & 83.0 \\
UniAD~\cite{uniad}       & C   & 97.8  & 91.9  & 78.8 & 92.9  & \textbf{100.0} & 83.4 \\
LTF~\cite{transfuser}           & C   & 97.4  & 92.8  & 79.0 & 92.4  & \textbf{100.0} & 83.8 \\
PARA-Drive~\cite{paradrive} & C & 97.9  & 92.4  & 79.3 & 93.0  & 99.8  & 84.0 \\
TransFuser~\cite{transfuser} & C+L & 97.7  & 92.8  & 79.2 & 92.8  & \textbf{100.0} & 84.0 \\
LAW~\cite{law}           & C   & 96.4  & 95.4  & 81.7 & 88.7  & 99.9  & 84.6 \\
DRAMA~\cite{drama}       & C+L & 98.0  & 93.1  & 80.1 & 94.8  & \textbf{100.0} & 85.5 \\
Hydra-MDP~\cite{li2024hydra} & C+L & 98.3  & 96.0  & 78.7 & 94.6  & \textbf{100.0} & 86.5 \\
WoTE~\cite{wote}         & C+L & \textbf{98.5}  & \textbf{96.8}  & \textbf{81.9} & 94.9  & 99.9  & 88.3 \\
\midrule
\rowcolor{gray!15}
\textbf{\method\ (rule)} & C+L & \textbf{98.5} & \textbf{96.8} & 81.7 & \textbf{97.7} & \textbf{100.0} & \textbf{89.4} \\
\rowcolor{gray!15}
\textbf{\method\ (learned)}     & C+L & 98.3 & 96.5 & 81.6 & 97.5 & \textbf{100.0} & 89.1 \\
\bottomrule
\end{tabular}
\end{table*}

\begin{table}[t]
\centering
\caption{Cost-map source ablation on NAVSIM \texttt{navtest}. The Cost-map source column reports how the per-cell cost map is constructed: a hand-designed rule cost, a rule base plus learned residual correction, or the full learned decoder. $N_2$, $N_3$, and $N_4$ are all \method\ configurations.}
\label{tab:b2_abl_source}
\small
\setlength{\tabcolsep}{6pt}
\renewcommand{\arraystretch}{1.05}
\begin{tabular*}{\columnwidth}{@{\hspace{3pt}\extracolsep{\fill}} l l c c @{\hspace{3pt}}}
\toprule
Method & Cost-map source & PDMS$\uparrow$ & TTC$\uparrow$ \\
\midrule
$N_1$              & --              & $88.29$          & $94.9$ \\
\midrule
$\boldsymbol{N_2}$ & Rule            & $\mathbf{89.36}$ & $\mathbf{97.7}$ \\
$N_3$              & Rule + Learned  & $88.88$          & $97.5$ \\
$N_4$              & Learned         & $89.07$          & $97.5$ \\
\bottomrule
\end{tabular*}
\end{table}

\begin{table}[t]
\centering
\caption{Training-signal ablation within the learned cost-map configuration on NAVSIM \texttt{navtest}. $\mathcal{L}_{\mathrm{cost}}$: BCE supervision against the host semantic mask; Style: dual AdaFiLM style branch with learned style codes.}
\label{tab:b2_abl_training}
\small
\setlength{\tabcolsep}{6pt}
\renewcommand{\arraystretch}{1.05}
\begin{tabular*}{\columnwidth}{@{\hspace{3pt}\extracolsep{\fill}} l c c c c @{\hspace{3pt}}}
\toprule
Method & $\mathcal{L}_{\mathrm{cost}}$ & Style & PDMS$\uparrow$ & TTC$\uparrow$ \\
\midrule
$N_1$              & --           & $\checkmark$ & $89.11$ & $97.4$ \\
$N_2$              & $\checkmark$ & --           & $89.14$ & $97.5$ \\
\midrule
\textbf{\method}   & $\checkmark$ & $\checkmark$ & $89.07$ & $97.5$ \\
\bottomrule
\end{tabular*}
\end{table}

\begin{table}[t]
\centering
\caption{PDMS stratified by WoTE baseline difficulty on NAVSIM \texttt{navtest}. $\Delta = \text{Learned} - \text{Rule}$.}
\label{tab:b2_difficulty}
\footnotesize
\setlength{\tabcolsep}{4pt}
\renewcommand{\arraystretch}{1.05}
\begin{tabular*}{\columnwidth}{@{\hspace{3pt}\extracolsep{\fill}} l c c c c c @{\hspace{3pt}}}
\toprule
Difficulty band & $N$ & WoTE & Rule & Learned & $\Delta$ \\
\midrule
Hard (PDMS$<0.7$)      &    990 & $24.18$ & $39.27$ & $\mathbf{56.47}$ & $\mathbf{+17.20}$ \\
Medium (0.7--0.9)  & 2\,961 & $85.74$ & $\mathbf{85.89}$ & $83.63$          & $-1.99$ \\
Easy ($\ge 0.9$)   & 8\,196 & $\mathbf{96.95}$ & $96.66$ & $94.62$          & $-2.42$ \\
\bottomrule
\end{tabular*}
\end{table}

\textit{(2) Ablation study.}\label{sec:b2_abl}
We organize the anchor-score ablation in two parts. Table~\ref{tab:b2_abl_source} compares cost-map sources (no module, rule, residual, learned), and Table~\ref{tab:b2_abl_training} ablates two training signals within the full learned configuration. Both follow the same evaluation protocol as the main results.

\textit{Cost-map source ablation (part a).} Table~\ref{tab:b2_abl_source} shows that the cost pathway drives the NAVSIM gain. All cost-map variants improve over $N_1$ on PDMS and TTC. The TTC gain is consistent across the three variants, from $+2.6$ to $+2.8$ points. This suggests that the reward-level coupling is effective once a reasonable spatial cost is available. It down-weights anchors that pass through high-cost cells, especially near dynamic obstacles.

Among the cost-map sources, the rule cost obtains the highest aggregate PDMS ($89.36$). The learned cost also improves over WoTE, reaching $89.07$. The residual form is weaker ($88.88$). This indicates that rule and learned costs should not be merged by a simple pointwise addition. The two costs encode different preferences, and their sum can disturb the anchor ranking instead of producing a clearer reward.

\textit{Training-signal ablation (part b).} Table~\ref{tab:b2_abl_training} shows that removing cost-map supervision or style conditioning does not hurt aggregate PDMS. Both variants are slightly above the full learned model. All rows still keep the cost-map decoder and receive planning gradients through the fused reward, so the result mainly indicates that $\mathcal{L}_{\mathrm{cost}}$ and style conditioning are weakly reflected by the NAVSIM aggregate score. We therefore interpret the learned cost map beyond aggregate PDMS: Sec.~\ref{sec:b2_strata} examines where it helps across scene difficulty, and Sec.~\ref{sec:qual_style} examines whether style conditioning changes the intermediate cost map as intended.

\textit{(3) Scene-level analysis.}\label{sec:b2_strata}
Throughout this section, \emph{Rule} and \emph{Learned} refer to $N_2$ and $N_4$ in Table~\ref{tab:b2_abl_source}. The difficulty split is used only for analysis, not for training or model selection. 

Since easy scenes dominate \texttt{navtest}, aggregate PDMS can hide where the learned cost map helps. The 4\,316 scenes on which WoTE passes every sub-metric contribute $35.5\%$ of the benchmark weight, with a cross-variant PDMS spread below $2.1$. We therefore use two baseline-anchored diagnostics: the WoTE-PDMS difficulty bands in Table~\ref{tab:b2_difficulty} and the compositional difficulty count defined as
\begin{equation}
n_{\mathrm{chal}}(\mathrm{scene}) = \left|\left\{ k \in \mathcal{S} : m_k^{\mathrm{WoTE}}(\mathrm{scene}) < 0.95 \right\}\right|,
\end{equation}
where $\mathcal{S} = \{\mathrm{NC}, \mathrm{DAC}, \mathrm{EP}, \mathrm{TTC}, \mathrm{Cmf}\}$ and $m_k^{\mathrm{WoTE}}(\mathrm{scene})$ is the normalized per-scene value of metric $k$ produced by the reproduced WoTE baseline. Both diagnostics depend only on the baseline evaluation.

\begin{figure}[t]
\centering
\includegraphics[width=\columnwidth]{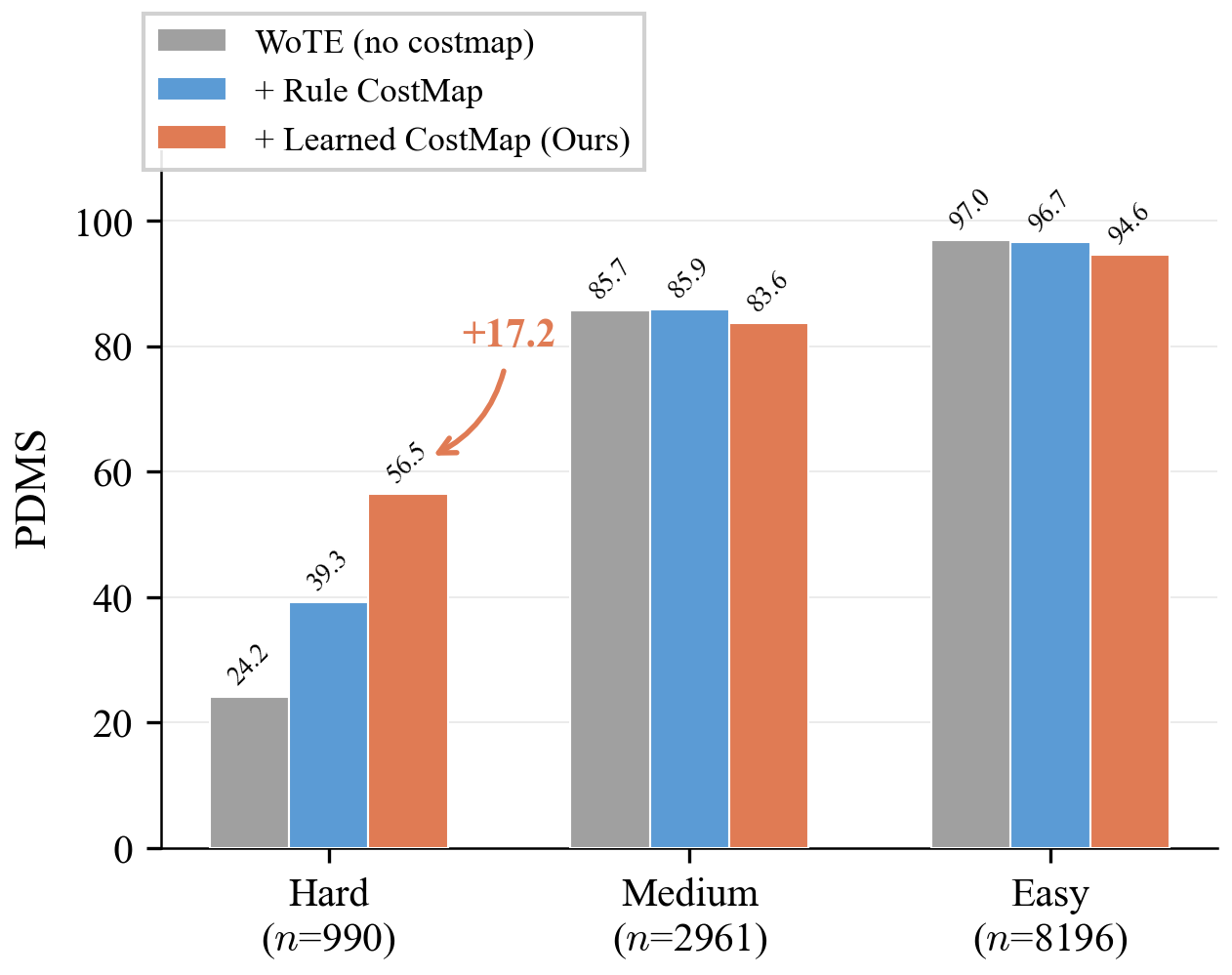}
\caption{PDMS stratified by reproduced WoTE baseline difficulty on NAVSIM \texttt{navtest}. The learned cost map achieves its largest gain ($+17.2$) on hard scenes where the baseline and rule-based alternatives score lower, while the rule variant remains stronger on medium and easy cases.}
\label{fig:difficulty}
\end{figure}

Table~\ref{tab:b2_difficulty} and Fig.~\ref{fig:difficulty} show that the relative performance of the learned and rule-based cost maps depends on scene difficulty. The learned cost map is stronger on hard scenes, with a $+17.20$ PDMS gain over the rule, while the rule remains stronger on medium and easy scenes. Since easy scenes are more than eight times more frequent than hard scenes, the aggregate score still favors the rule. This split indicates different operating points rather than universal superiority of either cost.

\begin{table}[t]
\centering
\caption{Sub-metric breakdown on the hard subset ($N=990$) on NAVSIM \texttt{navtest}. $\Delta = \text{Learned} - \text{Rule}$. The PDMS row is computed from the sub-metric averages above.}
\label{tab:b2_submetric}
\footnotesize
\setlength{\tabcolsep}{4pt}
\renewcommand{\arraystretch}{1.05}
\begin{tabular*}{\columnwidth}{@{\hspace{3pt}\extracolsep{\fill}} l c c c c @{\hspace{3pt}}}
\toprule
Metric & WoTE & Rule & Learned & $\Delta$ \\
\midrule
No-at-fault coll.          & $73.0$  & $72.1$  & $\mathbf{78.9}$  & $+6.7$  \\
Drivable compliance        & $42.4$  & $41.4$  & $\mathbf{67.2}$  & $+25.8$ \\
Ego progress               & $15.0$  & $13.0$  & $\mathbf{40.3}$  & $+27.3$ \\
Time-to-collision          & $57.7$  & $58.3$  & $\mathbf{72.1}$  & $+13.8$ \\
Comfort                    & $\mathbf{100.0}$ & $\mathbf{100.0}$ & $\mathbf{100.0}$          & $0.0$   \\
Direction comp.            & $\mathbf{100.0}$ & $91.8$  & $93.2$           & $+1.4$  \\
\midrule
PDMS                       & $10.31$ & $8.31$  & $\mathbf{41.40}$ & $+33.08$ \\
\bottomrule
\end{tabular*}
\end{table}

Table~\ref{tab:b2_submetric} explains the hard-scene gain. The learned cost map improves drivable-area compliance, ego progress, and TTC by $+25.8$, $+27.3$, and $+13.8$, respectively. The ego-progress gain shows that the learned variant does not improve hard scenes by simply braking. It maintains progress while also improving safety-related metrics, which is difficult for a uniform rule cost with fixed penalties across scene contexts. The PDMS row is recomputed from the displayed sub-metric averages and is used only to summarize this subset.

The scene-level win rate across challenge counts further supports this difficulty-dependent interpretation. With ties excluded, the learned variant wins over the rule in $30.5\%$, $55.4\%$, $71.0\%$, $87.9\%$, and $100\%$ of decided cases as $n_{\mathrm{chal}}$ increases from 0 to 4. The learned cost map is therefore most useful when several PDMS requirements fail at the same time. Its lower aggregate score is mainly a weighting effect caused by the easy-scene majority.

\textit{Oracle complementarity.}\label{sec:b2_oracle}
Table~\ref{tab:b2_oracle} reports an oracle diagnostic that selects, for each scene, whichever variant obtains the higher PDMS. The oracle reaches $91.78$ PDMS, which is $+2.42$ above the best single variant. A coarse difficulty-band diagnostic using Learned on hard scenes and Rule otherwise would reach about $90.76$ PDMS, or $+1.40$ over the rule alone. These evaluation-time diagnostics further show that the learned and rule-based costs are useful in different parts of the scene distribution.

\begin{table}[t]
\centering
\caption{Oracle complementarity on NAVSIM \texttt{navtest}. Oracle selects, per scene, whichever variant (Rule or Learned) attains the higher PDMS.}
\label{tab:b2_oracle}
\small
\setlength{\tabcolsep}{6pt}
\renewcommand{\arraystretch}{1.05}
\begin{tabular*}{\columnwidth}{@{\hspace{3pt}\extracolsep{\fill}} l c @{\hspace{3pt}}}
\toprule
Method & PDMS$\uparrow$ \\
\midrule
Rule                              & $89.36$ \\
Learned                           & $89.07$ \\
\midrule
\textbf{Oracle (best per scene)}  & $\mathbf{91.78}$ \\
\bottomrule
\end{tabular*}
\end{table}

\textit{Summary across planner families.} The NAVSIM results complement the nuScenes results. Together, they show that the same cost-map abstraction can be coupled with both a regression planner and an anchor-score planner. This supports the portability of the proposed bridge across representative planner families. The two instantiations still use host-specific adapters, resolutions, auxiliary targets, and independently trained weights, so the results support representative portability rather than universal host-agnostic deployment.

\subsection{Qualitative Analysis}\label{sec:qual_costmap}
\begin{figure}[t]
\centering
\includegraphics[width=\columnwidth]{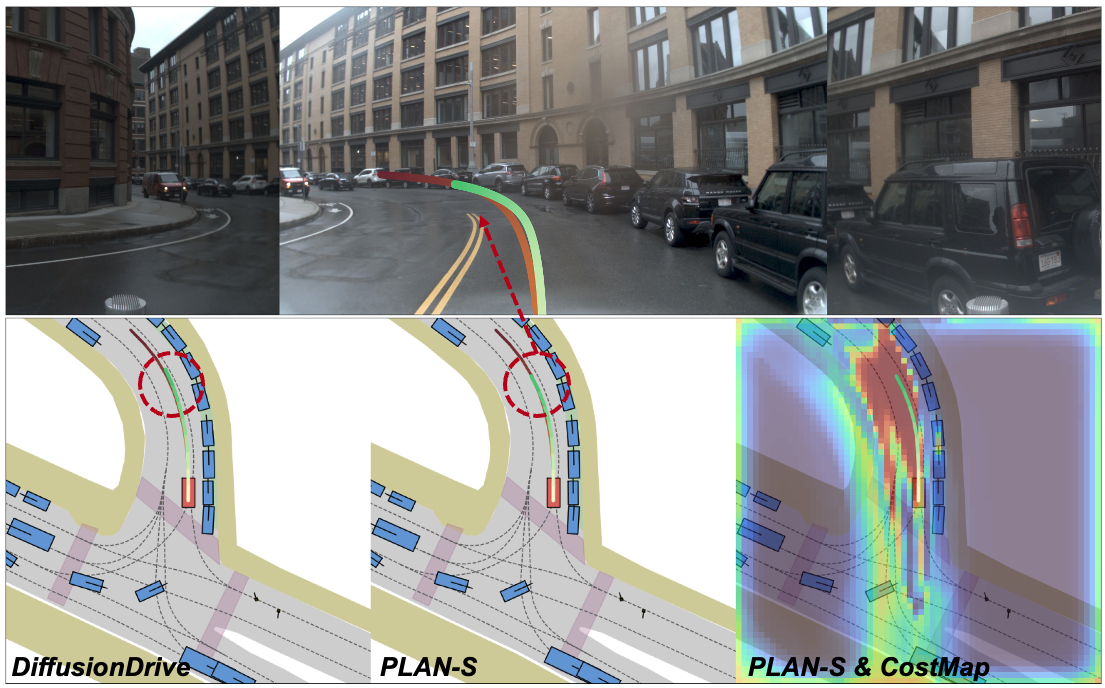}
\caption{Qualitative comparison on a NAVSIM \texttt{navtest} scene. \textbf{Top}: multi-view inputs with the front-view trajectory overlay (green: model prediction, red: GT). \textbf{Bottom}: BEV outputs for DiffusionDrive, \method, and \method with the learned cost map overlaid; red tones denote lower-cost regions, and blue tones denote higher-cost regions.}
\label{fig:qualitative}
\end{figure}

Fig.~\ref{fig:qualitative} shows a representative NAVSIM \texttt{navtest} scene at a curved urban intersection. The top row provides the multi-view camera inputs and the front-view trajectory overlay, where green denotes the model prediction and red denotes the GT trajectory. The bottom row compares BEV planning outputs. Compared with DiffusionDrive, \method keeps a smoother predicted trajectory with larger clearance from nearby agents. The rightmost panel overlays the learned four-channel cost map, where blue regions appear around dynamic obstacles and off-road boundaries, while red regions remain along the lower-cost drivable corridor.

This visual pattern is consistent with the quantitative results. The learned cost map does not simply shift the route. Instead, it marks local regions where the planner should be more cautious. This explains why the main gains appear in safety-related metrics, including the TTC gains in Table~\ref{tab:b2_abl_source} and the CR reduction in Table~\ref{tab:b1_main}. It also supports the role of the cost map as an explicit spatial representation that can be inspected before the final trajectory is selected.

\begin{figure}[t]
\centering
\includegraphics[width=\columnwidth]{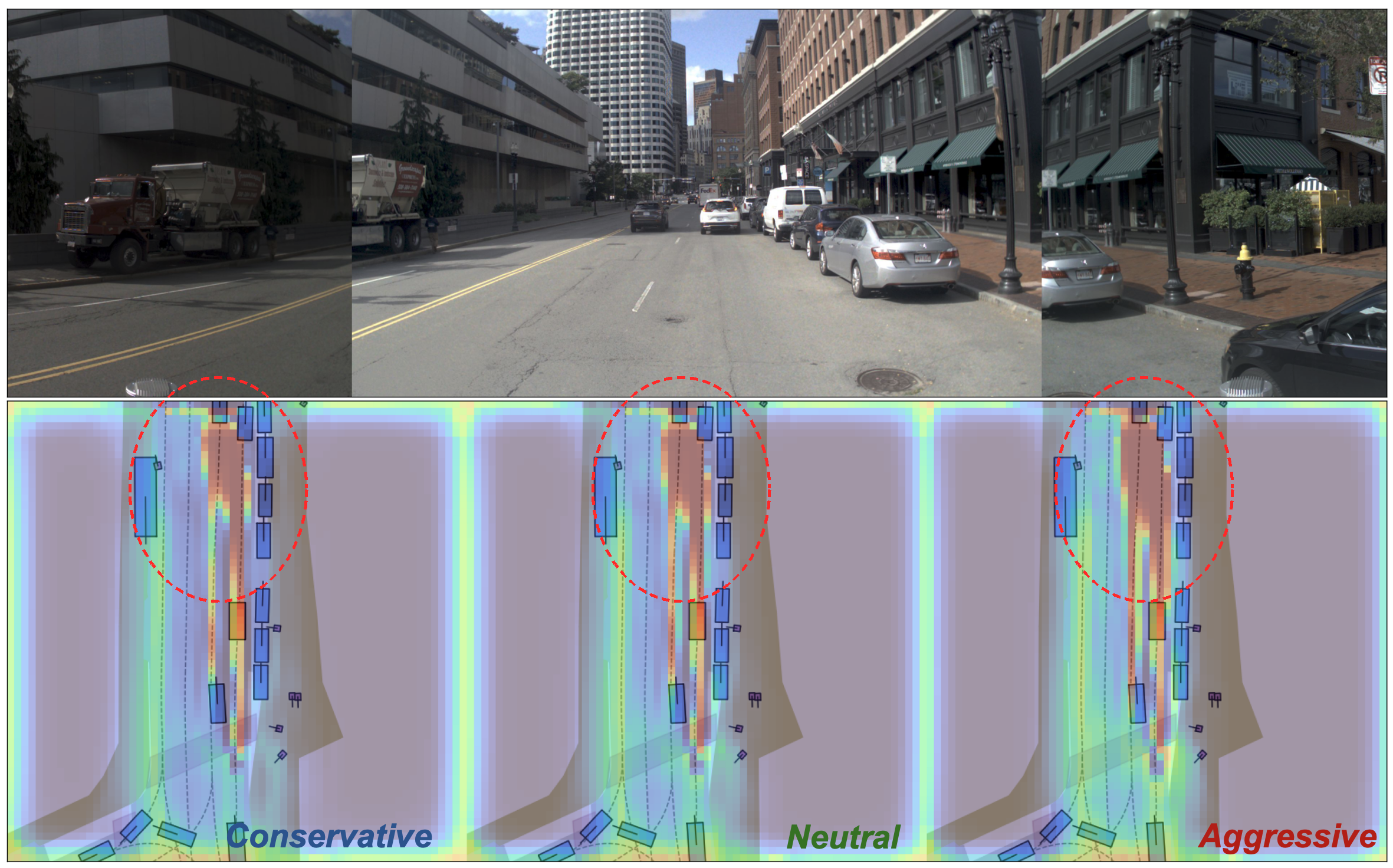}
\caption{Style-conditioned cost-map comparison on a NAVSIM \texttt{navtest} scene. \textbf{Top}: front-view scene context. \textbf{Bottom}: BEV cost-map overlays under conservative, neutral, and aggressive style codes. Red dashed circles highlight regions where the spatial preference encoded by the cost map changes visibly across driving styles.}
\label{fig:style}
\end{figure}

\textit{Style-conditioned cost maps.}\label{sec:qual_style}
Fig.~\ref{fig:style} illustrates style-conditioned cost maps on a straight urban road with parked vehicles and a truck in the adjacent lane. For this controlled visualization, conservative, neutral, and aggressive correspond to three preset two-dimensional style codes, $(0.0,0.0)$, $(0.5,0.5)$, and $(1.0,1.0)$, respectively, rather than external human labels. The top row provides the front-view scene context. The bottom row shows the BEV cost maps generated under the three style settings. Conservative intent expands the blue high-cost regions near surrounding vehicles. Aggressive intent contracts these regions, and the neutral setting lies between them.

The cost-map changes are concentrated around nearby vehicles and lane-adjacent regions, while the lower-cost drivable corridor remains aligned with the ego lane across all three styles. \method therefore produces diverse cost maps, with spatially consistent variations aligned to different driving styles, while the route semantics remain unchanged. Together with Table~\ref{tab:b2_abl_training}, this result suggests that style conditioning is better understood as a controllability mechanism for the intermediate cost representation, not as a direct way to maximize aggregate PDMS.

Aggregate PDMS is therefore insufficient for evaluating this aspect of \method. It measures general closed-loop driving quality, but it does not assess whether the cost map follows the requested driving style or whether the intermediate representation is interpretable. A complete style evaluation should combine aggregate driving metrics with style-aware preference metrics, such as SM-PDMS~\cite{styledrive}, and direct checks of the intermediate cost map.

\section{Discussion}\label{sec:discussion}

\textbf{Evidence boundary and empirical interpretation.} The present experiments support representative architectural portability rather than universal host-agnostic deployment. \method reuses the same cost-map abstraction, channel semantics, and prior-plus-refinement principle across one regression host and one anchor-score host, while host-side adapters, resolutions, auxiliary targets, and learned weights remain instantiation-specific. On nuScenes, the cost pathway contributes most clearly to collision reduction, while the L2 change is comparatively modest and is best viewed as a secondary effect.
On NAVSIM, the rule variant attains the highest aggregate score ($89.4$ PDMS), confirming that a fixed hand-designed cost remains a strong default on routine scenes. The learned variant is better characterized as a context-sensitive, style-conditioned bridge whose hard-scene benefit is partly obscured by aggregate scoring.

\textbf{Hard-scene behavior and complementarity.} The hard-band gain of the learned variant over the rule ($+17.2$ PDMS) reflects an imbalance present in driving datasets: routine scenes dominate, while hard situations such as tight gaps, dense pedestrian crossings, and interacting agents are rarer. A rule cost applies the same penalty structure regardless of scene complexity, whereas a learned head captures context-dependent regularities such as the spatial co-occurrence of dynamic agents and narrow drivable corridors that no fixed rule can express. The oracle result uses evaluation-time PDMS and is therefore an analysis tool rather than a deployable model; it quantifies the complementarity between the two cost designs.

\textbf{Style conditioning, metric limitations, and remaining gaps.} Dual AdaFiLM produces spatially consistent cost-map variations across driving intents (Sec.~\ref{sec:qual_style}), changing the cost distribution around nearby vehicles while preserving the lower-cost drivable corridor. The benchmarks used in this paper do not directly evaluate style matching: NAVSIM PDMS measures general driving quality, and StyleDrive-style metrics~\cite{styledrive} score final trajectory behavior against discrete human style categories, whereas \method modulates the upstream spatial cost representation. Such metrics evaluate final trajectory behavior and require a clear mapping between the style input, the target style label, and human driving behavior. They do not directly assess whether the intermediate cost map changes consistently with the requested style.
A complete style evaluation should combine general driving-quality scores, style-aware preference metrics, human preference labels, and intermediate-representation checks. Future work will extend the evaluation to additional hosts and multi-seed statistics, while further reducing host-specific auxiliary supervision.

\section{Conclusion}\label{sec:conclusion}
We presented \method, a planner-facing bridge that explicitly models latent style dynamics for LWM-based autonomous driving. \method decodes a style-conditioned, four-channel semantic cost map from the latent representation and consumes it before final trajectory selection. The same cost-map contract is instantiated through attention-level fusion for regression planning and reward-level fusion for anchor scoring. On the regression host (nuScenes), \method reduces L2 at every horizon over the baseline (with $0.55$\,m average L2) and yields a $42\%$ relative reduction in the 3\,s collision rate. The ablation study further shows that the cost pathway is most directly reflected in collision reduction. On the anchor-score host (NAVSIM), the rule-based cost obtains the highest aggregate score ($89.4$ PDMS), while the learned style-conditioned cost provides complementary benefits on baseline-challenging scenes. Qualitative results further show that \method can produce diverse cost maps, with spatially consistent variations aligned to different driving styles. These results support explicit latent style dynamics as a practical way to improve controllability and interpretability in LWM-based planning. Extending the evaluation to additional hosts, multi-seed statistics, and quantitative style metrics remains the principal direction for future work.

\section*{Acknowledgment}
This work was partially supported by National Key Research and Development Program of China 2024YFB4707603, NSFC under Grants 62373298, U24A20252, and 62373315, Guangdong Provincial Project 2023ZT10X009, Guangdong Provincial Key Lab of Integrated Communication, Sensing and Computation for Ubiquitous Internet of Things (No. 2023B1212010007), Nansha Key Science and Technology Project under Grant 2023ZD006, and Guangzhou Bureau of Education College Scientific Research Project 2024312169.
\bibliographystyle{IEEEtran}
\bibliography{references}

\begin{IEEEbiography}
[{\includegraphics[width=1in,height=1.25in,clip,keepaspectratio]{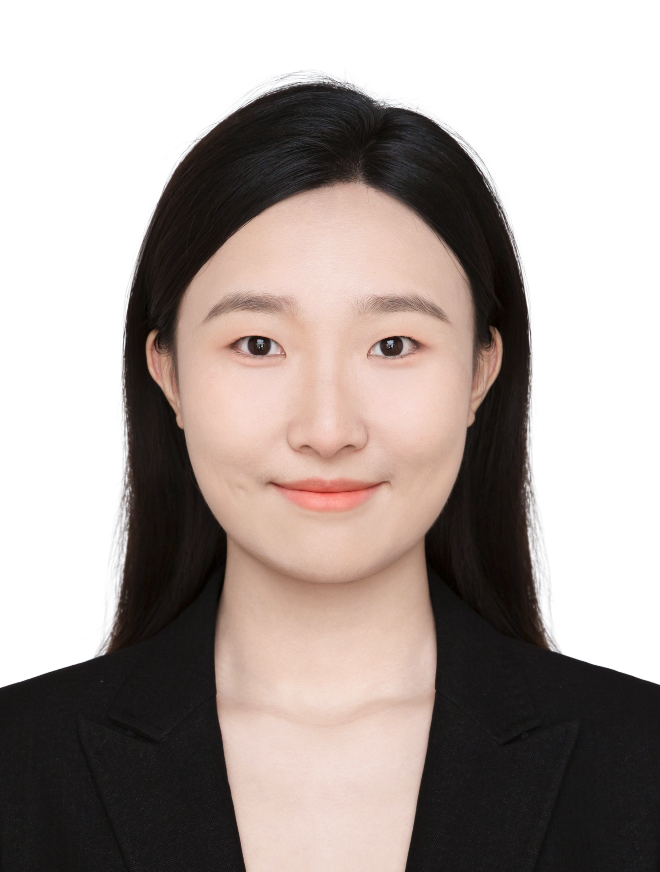}}]{Xiaoyun Qiu}
received the B.S. and M.S. degrees in transportation engineering from Harbin Institute of Technology, Harbin, China, in 2019 and 2021, respectively. She is currently working toward the Ph.D. degree in intelligent transportation from the Hong Kong University of Science and Technology (Guangzhou), Guangzhou, China.
Her research interests include autonomous driving decision-making, driving style recognition, and latent world models for autonomous driving.
\end{IEEEbiography}
\vspace{-2em}
\begin{IEEEbiography}
[{\includegraphics[width=1in,height=1.25in,clip,keepaspectratio]{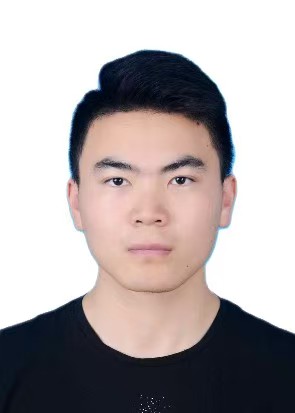}}]{Jingtao He} received the B.E. degree in Mechanical Engineering from Chongqing University in 2020, and the M.S. degree in Electronic and Information Engineering from Xi'an Jiaotong University in 2024. He is currently pursuing the Ph.D. degree at The Hong Kong University of Science and Technology (Guangzhou). His research interests include end-to-end autonomous driving, Vision-Language-Action (VLA) models, and multimodal large language models for intelligent transportation systems.
\end{IEEEbiography}
\begin{IEEEbiography}
[{\includegraphics[width=1in,height=1.25in,clip,keepaspectratio]{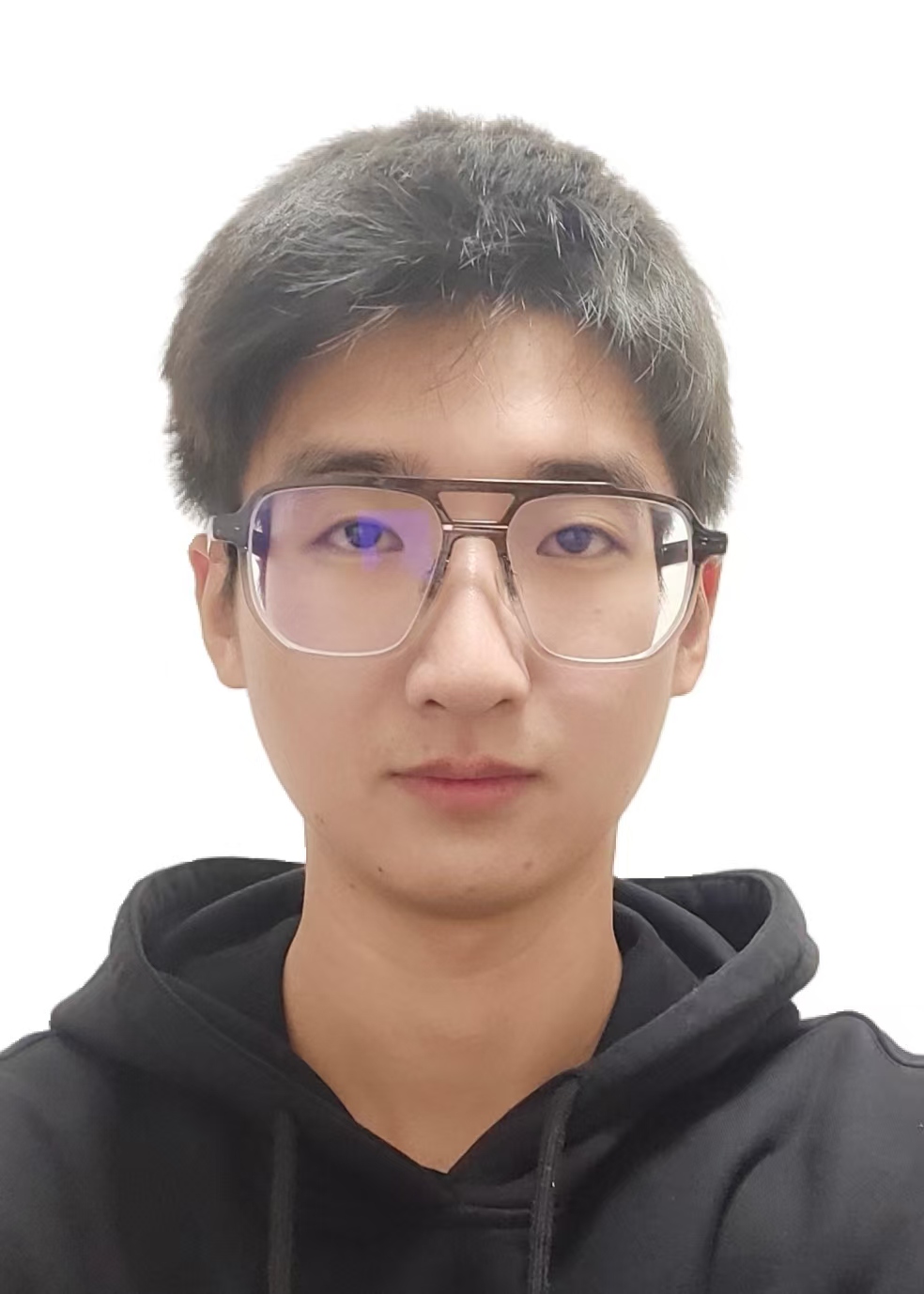}}]{Yijie Chen} received the B.Eng. degree in Traffic and Transportation from Beijing Jiaotong University in 2022. He received M.S. degree in Intelligent Transportation from the Hong Kong University of Science and Technology (Guangzhou) in 2024, where he is currently pursuing the Ph.D. degree with the Intelligent Transportation Thrust. His current research interests include autonomous driving, cooperative perception and prediction, and intelligent transportation systems.
\end{IEEEbiography}
\vspace{-1em}
\begin{IEEEbiography}
[{\includegraphics[width=1in,height=1.25in,clip,keepaspectratio]{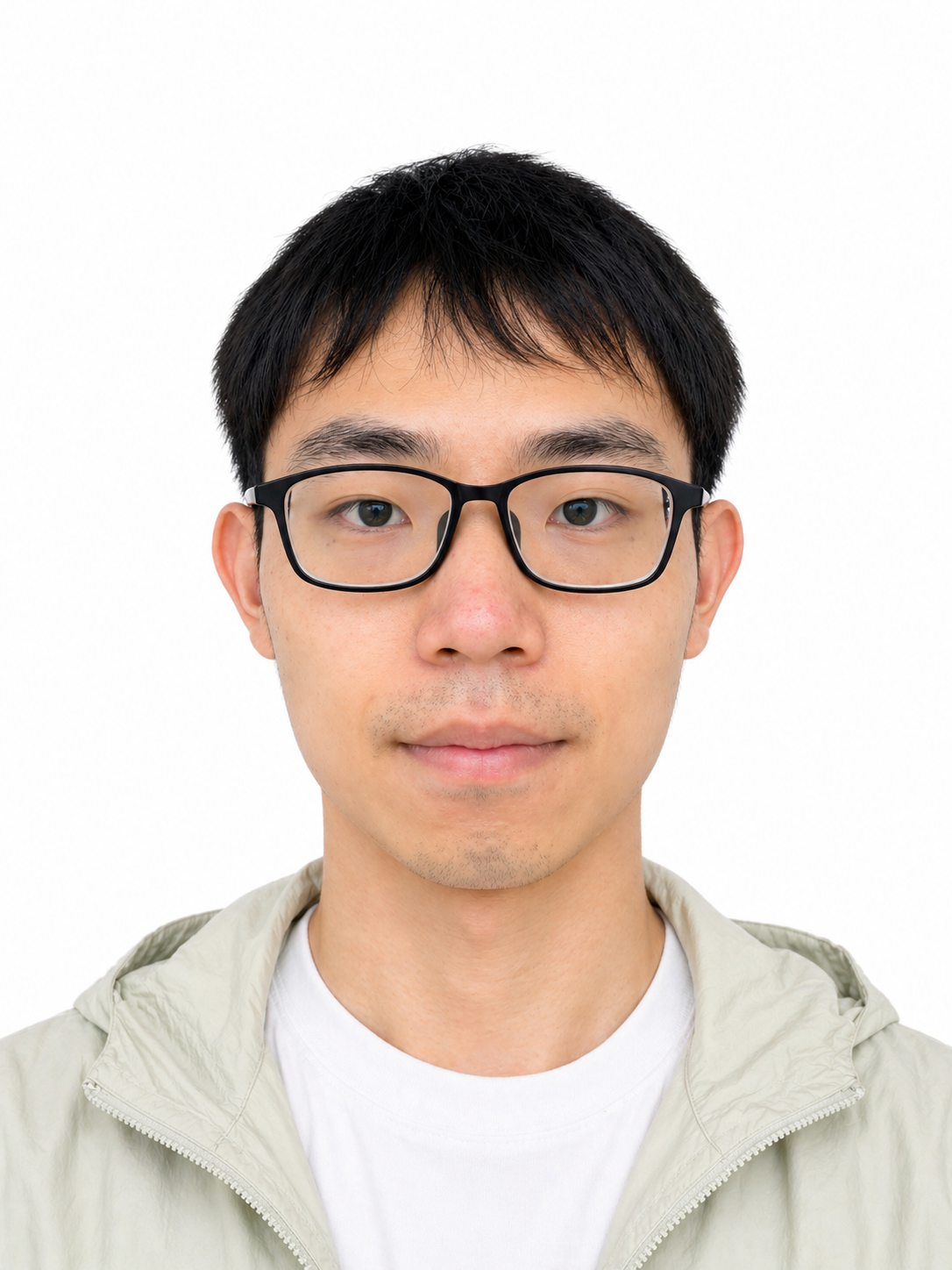}}]{Yusong Huang} received his bachelor’s degree from the School of Software Engineering, Beijing Jiaotong University, in 2025. He is currently a PhD student at The Hong Kong University of Science and Technology (Guangzhou). His research interests include world models and embodied intelligence.
\end{IEEEbiography}
\begin{IEEEbiography}
[{\includegraphics[width=1in,height=1.25in,clip,keepaspectratio]{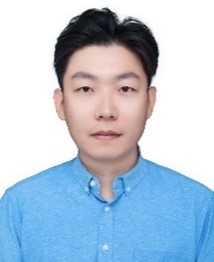}}]{Haotian Wang}
(Member, IEEE) received the Ph.D. degree from the Institute of Artificial Intelligence and Robotics, Xi’an Jiaotong University, Xi’an, China, in 2025. He was a visiting Ph.D. student at Nanyang Technological University, Singapore, from 2023 to 2024. He is currently a Postdoctoral Fellow with The Hong Kong University of Science and Technology (Guangzhou), Guangzhou, China, with an additional postdoctoral affiliation with The Chinese University of Hong Kong, Hong Kong, China. His research interests include spatial intelligence, 3D vision, multi-modal vision, and embodied intelligence.
\end{IEEEbiography}
\begin{IEEEbiography}
[{\includegraphics[width=1in,height=1.25in,clip,keepaspectratio]{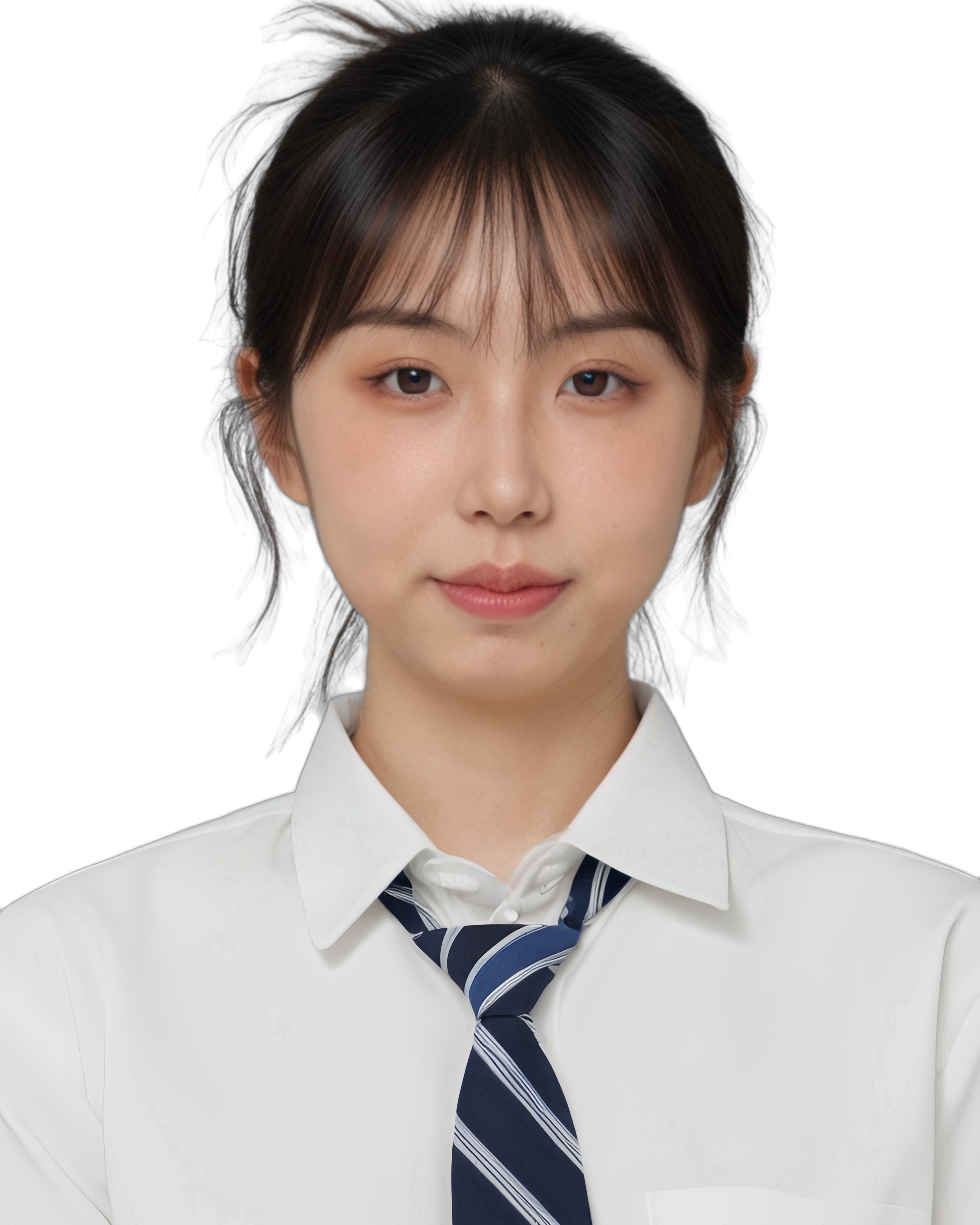}}]{Yixuan Wang} received the B.E. degree from Tianjin University in 2019. She is currently pursuing the M.Phil. degree at The Hong Kong University of Science and Technology (Guangzhou). Her research interests include end-to-end autonomous driving, vision-language navigation, and the applications of multimodal large language models in decision-making agents.
\end{IEEEbiography}

\begin{IEEEbiography}
[{\includegraphics[width=1in,height=1.25in,clip,keepaspectratio]{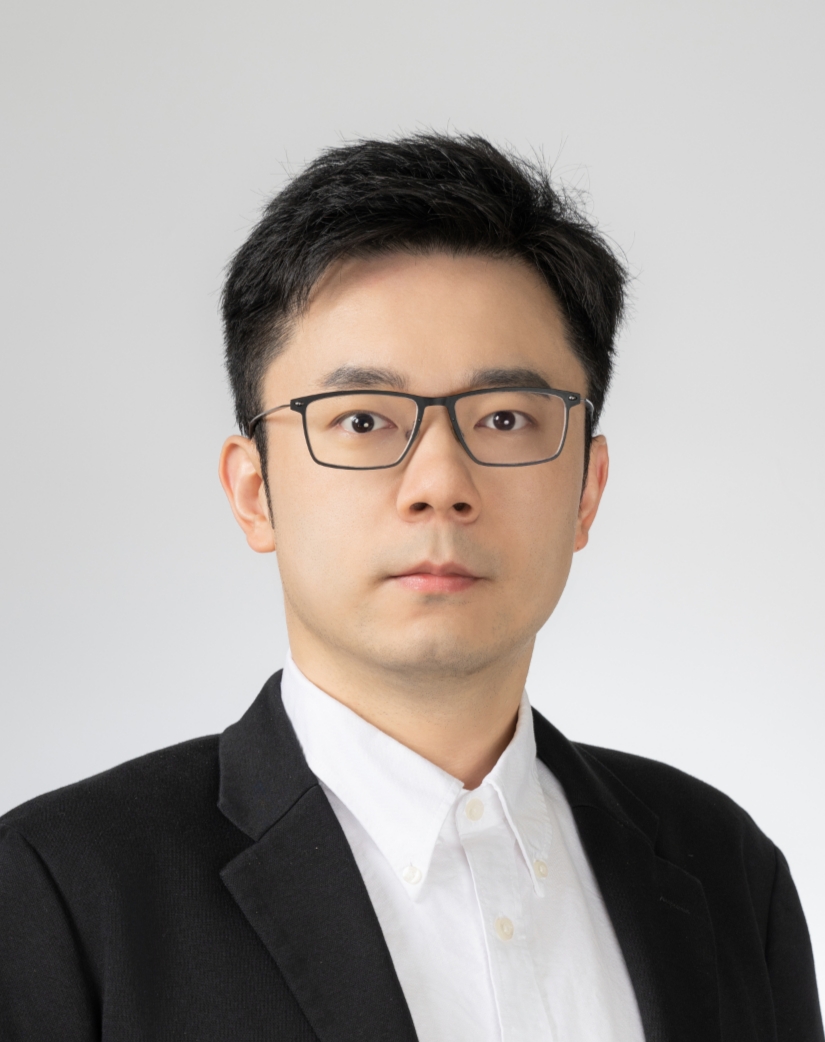}}]{Xinhu Zheng} is currently an Assistant Professor with the Intelligent Transportation Thrust of the Systems Hub, at Hong Kong University of Science and Technology (Guangzhou). He received the Ph.D. degree in Electrical and Computer Engineering from the University of Minnesota, Minneapolis. He has published more than 40 papers in peer-reviewed journals and conferences, including IEEE Transactions on Pattern Analysis and Machine Intelligence, IEEE Internet of Things Journal, IEEE Transactions on Intelligent Transportation Systems, IEEE Transactions on Intelligent Vehicles, etc. His current research interests include autonomous driving, connected intelligence, and embodied artificial intelligence. He is currently an associate editor for IEEE Transactions on Intelligent Vehicles and IEEE Transactions on Mobile Computing.
\end{IEEEbiography}

\end{document}